\documentclass[10pt,journal,compsoc]{IEEEtran}
\ifCLASSOPTIONcompsoc
  \usepackage[nocompress]{cite}
\else
  \usepackage{cite}
\fi

\usepackage{graphicx}
\usepackage{subfigure}
\usepackage{multirow}
\usepackage{amssymb}
\usepackage{amsmath}
\usepackage{booktabs}
\usepackage[ruled]{algorithm2e}
\usepackage{color}
\usepackage{ragged2e}
\usepackage{hyperref}
\usepackage{cite}

\ifCLASSINFOpdf
\else
\fi
\begin{document}
%
\title{Training Compact CNNs for Image Classification using Dynamic-coded Filter Fusion}
%
%
%
%

\author{Mingbao Lin,
        Bohong Chen,
        Fei Chao,~\IEEEmembership{Member,~IEEE},
        Rongrong Ji,~\IEEEmembership{Senior Member,~IEEE}

\IEEEcompsocitemizethanks{


\IEEEcompsocthanksitem M. Lin, B. Chen, F. Chao and R. Ji are with the Media Analytics and Computing Laboratory, Department of Artificial Intelligence, School of Informatics, Xiamen University, Xiamen 361005, China.\protect
\IEEEcompsocthanksitem M. Lin is also with the Tencent Youtu Lab, Shanghai 200233, China.\protect
\IEEEcompsocthanksitem R. Ji is also with the Instititue of Artificial Intelligence, Xiamen University, Xiamen 361005, China.\protect
}

\thanks{Manuscript received April 19, 2005; revised August 26, 2015.}}

%
%

\markboth{IEEE TRANSACTIONS ON PATTERN ANALYSIS AND MACHINE INTELLIGENCE Under Review}
{Shell \MakeLowercase{\textit{et al.}}: Bare Demo of IEEEtran.cls for Computer Society Journals}
%



\IEEEtitleabstractindextext{%
\begin{abstract}
\justifying
The mainstream approach for filter pruning is usually either to force a hard-coded importance estimation upon a computation-heavy pretrained model to select ``important'' filters, or to impose a hyperparameter-sensitive sparse constraint on the loss objective to regularize the network training. In this paper, we present a novel filter pruning method, dubbed dynamic-coded filter fusion (DCFF), to derive compact CNNs in a computation-economical and regularization-free manner for efficient image classification. Each filter in our DCFF is firstly given an inter-similarity distribution with a temperature parameter as a filter proxy, on top of which, a fresh Kullback-Leibler divergence based dynamic-coded criterion is proposed to evaluate the filter importance. In contrast to simply keeping high-score filters in other methods, we propose the concept of filter fusion, \textit{i.e.}, the weighted averages using the assigned proxies, as our preserved filters. We obtain a one-hot inter-similarity distribution as the temperature parameter approaches infinity. Thus, the relative importance of each filter can vary along with the training of the compact CNN, leading to dynamically changeable fused filters without both the dependency on the pretrained model and the introduction of sparse constraints. Extensive experiments on classification benchmarks demonstrate the superiority of our DCFF over the compared counterparts. For example, our DCFF derives a compact VGGNet-16 with only 72.77M FLOPs and 1.06M parameters while reaching top-1 accuracy of 93.47\% on CIFAR-10. A compact ResNet-50 is obtained with 63.8\% FLOPs and 58.6\% parameter reductions, retaining 75.60\% top-1 accuracy on ILSVRC-2012. Our code, narrower models and training logs are available at \url{https://github.com/lmbxmu/DCFF}.

\end{abstract}

}

\maketitle

\IEEEdisplaynontitleabstractindextext

%
\IEEEpeerreviewmaketitle

\IEEEraisesectionheading{\section{Introduction}\label{introduction}}

%
%
%
%
\IEEEPARstart
{C}{onvolutional} neural networks (CNNs) have revolutionized many visual tasks by enabling unprecedented performance, ranging from image classification~\cite{szegedy2015going,he2016deep}, object detection~\cite{ren2015faster,ramanathan2020dlwl}, visual tracking~\cite{nam2016learning,li2020joint} and many others. However, such a performance boost is often built on the basis of huge computation cost and increasing parameter amount. While it is possible to run a large-scale CNN in an environment with powerful GPUs, it is still very challenging to deploy a large CNN model on resource-constrained mobile devices and embedded systems that demand a real-time response. Thus, finding out parameter and computation redundancy in CNNs has become an active research area in computer vision.

\begin{figure}[!t]
\begin{center}
\includegraphics[height=0.38\linewidth]{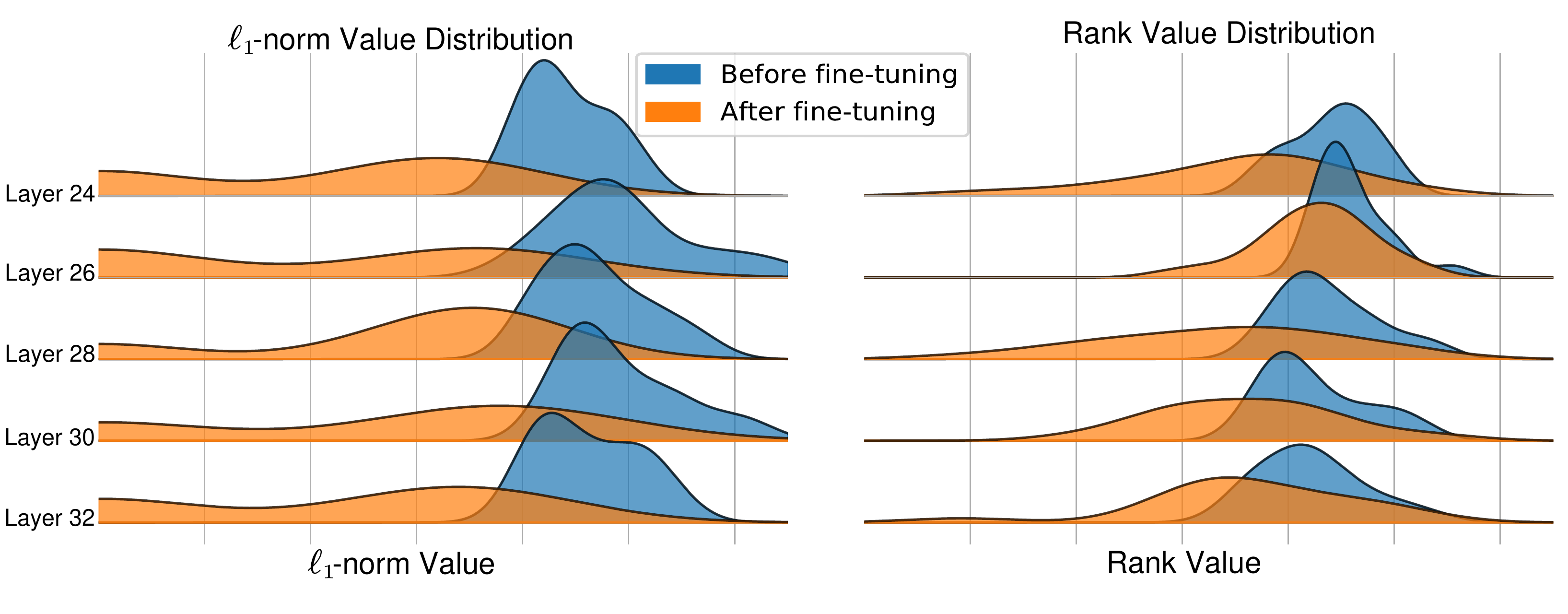}
\end{center}
\caption{\label{comparison}
\label{longtail}
Comparison of importance scores before and after fine-tuning. We select high-score filters using the criteria of $\ell_1$-norm~\cite{li2017pruning} and rank of feature map~\cite{lin2020hrank} from a pretrained ResNet-56. It can be observed that filters with high values of $\ell_1$-norm and rank of feature map have smaller values after fine-tuning.
}
\end{figure}
%
To this end, a large collection of research work has been spurred to derive compact CNNs, so as to improve the inference efficiency without the compromise on accuracy performance. Prevailing methods include, but are not limited to, weight sharing~\cite{chen2015compressing,gong2015compressing,zhu2017trained}, low-precision quantization~\cite{lin2020rotated,yang2019quantization,qu2020adaptive}, tensor decomposition~\cite{denil2013predicting,jaderberg2014speeding,kim2019efficient}, knowledge distillation~\cite{polino2018model,li2020gan,guo2020online} and network pruning~\cite{lin2021filter,lin2021network,guo2020multi}. 

Among these methods, pruning convolutional filters, \emph{a.k.a.} filter pruning, has attracted increasing attention since it removes entire filters without changing the original convolution structures and thus without extra requirements for inference engines. According to its procedures of learning compact CNNs, we generally categorize existing methods into \textit{pretraining-dependency} filter pruning and \textit{regularized-retraining} filter pruning.

\begin{figure*}[!t]
\begin{center}
\includegraphics[height=0.32\linewidth]{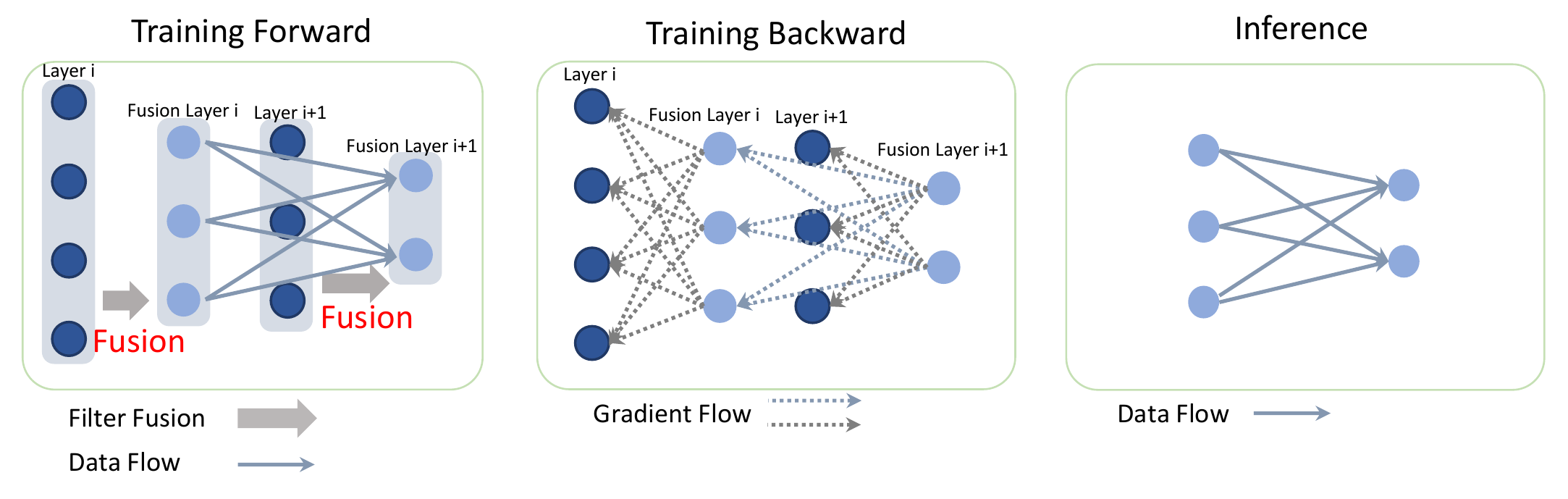}
\end{center}
\caption{\label{flow} Training and inference flows of our dynamic-coded filter fusion. Training: In the forward step, the original filters in the $i$-th layer are fused into a smaller group of filters that form the $i$-th fusion layer. The fusion layers make up the network backbone to process the input images. Notice the fused filters are intermediate results. They are unlearnable and provide gradients to update the original filters according to the chain rule.
Inference: Since the inference only involves forward propagation, the fused filters are preserved and serve as our compact model for an efficient inference.
}
\end{figure*}

\textbf{Pretraining-dependency}. A bunch of existing methods build filter pruning on top of a pretrained CNN model~\cite{hu2016network,molchanov2017pruning,he2017channel,li2017pruning,luo2017thinet,yu2018nisp,dubey2018coreset,lin2020hrank,luo2020neural}. To that effect, many studies aim to preserve ``important'' filters measured by an intrinsic criterion based on either pretrained filter weights such as ${\ell}_1$-norm~\cite{li2017pruning} and coreset~\cite{dubey2018coreset}, or data-driven activations such as output sparsity~\cite{hu2016network}, rank of feature map~\cite{lin2020hrank} and influence to the accuracy or loss~\cite{molchanov2017pruning,luo2020neural}. Another group formulates filter pruning as an iterative optimization problem to minimize reconstruction errors~\cite{he2017channel,luo2017thinet,yu2018nisp}. However, for all these methods, the capacity of pruned CNNs seriously relies on a computation-heavy pretrained model. Besides, the filter selection is hard-coded where the ``important'' filters are fixed, incurring a bottleneck of performance improvement~\cite{he2018soft}. In particular, fine-tuning is required to boost the accuracy. However, such fine-tuning is even more expensive than pretraining a large-scale CNN when implemented in layer-wise fashion~\cite{he2017channel,luo2017thinet,lin2020hrank}. As illustrated in Fig.\,\ref{comparison}, the ``important'' filters using $\ell_1$-norm~\cite{li2017pruning} or rank of feature map~\cite{lin2020hrank} no longer maintain high scores after fine-tuning. This phenomenon contradicts the motivation that high-score filters are more important, implying that these criteria could not capture the filter importance exactly.

\textbf{Regularized-retraining}. 
This direction embeds hand-crafted regularization rules into the network training loss~\cite{liu2017learning,huang2018data,zhao2019variational,li2020group,kang2020operation,luo2020autopruner,li2019oicsr,ding2020lossless}. To this end, the introduced regularization is typically adopted to sparsify a particular target, such as parameters of the batch normalization layer~\cite{liu2017learning,zhao2019variational,kang2020operation}, channel-level masks~\cite{huang2018data,luo2020autopruner}, auxiliary matrix~\cite{li2020group} and filter weights~\cite{li2019oicsr,ding2020lossless}. These studies employ a joint-retraining optimization and then a compact CNN model is obtained through preserving the retrained filters with large values. Although this strategy removes the dependency on a pretrained model, it also poses a great difficulty to the universality and flexibility of the training loss since the introduced sparse constraint is hyperparameter-sensitive and usually requires several rounds of complicated analyses. Moreover, some of these methods suffer the hurdle in optimizing this modified loss when training deep neural networks using common Stochastic Gradient Descent (SGD), and thus, these methods require specialized optimizers~\cite{huang2018data} and even another round of fine-tuning to boost the performance~\cite{lin2020channel}, both of which greatly affect the flexibility and ease of using these methods.

Overall, training compact CNNs through filter pruning remains an open question so far. The practical deployment requires not only more compact CNNs with high performance, but also a simple implementation. To this end, in this paper, we present a novel method for training compact CNNs, dubbed dynamic-coded filter fusion (DCFF), which removes the dependency on pretraining a large-scale CNN model and the introduction of sparse constraints. We first explore an inter-similarity among all filters and develop a distribution proxy with a temperature parameter for each filter, based on which we measure the importance of each filter via calculating the difference between its proxy and others, characterized by the Kullback-Leibler divergence. Unlike the pretraining-dependency studies that implement compact models in a hard-coded manner, we conduct the filter pruning in a dynamic-coded manner, where the distribution proxy degenerates to a one-hot distribution as the temperature parameter approaches infinity. Thus the relative importance of each filter can be dynamically captured along with the training of our compact CNNs. Then, instead of simply discarding low-score filters and fine-tuning high-score ones, we propose to fuse all filters through the weighted average using the assigned proxy. Finally, we train a compact CNN model from scratch to remove the dependency on pretraining a large-scale CNN model, leading to a major reduction in processing time. As shown in Fig.\,\ref{flow}, in the forward step, we only use fused filters to process the input data while the original filters are allowed to update in the backward step. After a regular network training, we can preserve the fused filters for inference and thus our DCFF obtains compact CNNs without auxiliary sparsity constraints, which well facilitates its practical usage and greatly differentiates our method from off-the-shelf regularized-retraining studies.

We haved conducted extensive experiments on CIFAR-10~\cite{krizhevsky2009learning} using VGGNet-16~\cite{simonyan2015very}, GoogLeNet~\cite{szegedy2015going} and ResNet-56/110~\cite{he2016deep}, and on ILSVRC-2012~\cite{russakovsky2015imagenet} using ResNet-50~\cite{he2016deep} and MobiletNet-V2~\cite{sandler2018mobilenetv2}. The results demonstrate the superior performance of our DCFF over many the competitors. 

In summary, the main contributions we have made in this paper include:

\begin{itemize}
  \item By exploring the inter-similarity among filters, a fresh Kullback-Leibler divergence-based measure is developed to evaluate the filter importance, which can dynamically select important filters along with the training of the compact CNN without the dependency on pretraining a computation-heavy model.

  \item By utilizing the inter-similarity distribution, a novel concept of filter fusion is put forward to achieve the training of the compact CNN, which implements filter pruning without discarding the low-score filters, and eliminates the introduction of hyperparameter-sensitive sparsity constraints.

  \item Through extensive experimental verification, our proposed DCFF not only advances in its simple implementation, but also shows a greater ability to reduce the model complexity over a variety of state-of-the-art studies, both of which increase the practical deployment of our method.
\end{itemize}

 

\section{Related Work}\label{related}

We discuss the major topics that are the most related to this paper. A more comprehensive overview can be found from the recent survey~\cite{vadera2020methods}.

\textbf{Weight Pruning}. 
Weight pruning removes individual neurons in filters or connections between fully-connected layers. Pioneers, optimal brain damage and optimal brain surgeon~\cite{lecun1990optimal,hassibi1993second}, utilize the second-order Hessian to prune weights. Han \emph{et al}.~\cite{han2015learning} proposed to recursively remove small-weight connectivity and retrain the $\ell_2$-regularized subnetwork to derive smaller 
weight values. Dynamic network surgery~\cite{guo2016dynamic} performs pruning and splicing on-the-fly, where the former compresses the network and the latter recovers the incorrect pruning. Aghasi \emph{et al}.~\cite{aghasi2017net} formulated the pruning as a convex optimization, which seeks per-layer sparse weights that maintain the inputs and outputs close to the original model. In~\cite{liu2018frequency}, 2-D DCT transformation is applied to sparsify the coefficients for spatial redundancy removal. The lottery ticket hypothesis~\cite{frankle2019lottery} randomly initializes a dense network and trains it from scratch. The subnets with high-weight values are extracted, and retrained with the initial weight values of the original dense model. Lin \emph{et al}.~\cite{lin2020dynamic} proposed a dynamic allocation of sparsity pattern and incorporated feedback signal to reactivate prematurely pruned weights. However, weight pruning results in an irregular sparsity which hardly supports practical speedup without delicate hardware/software~\cite{han2016eie}.

\textbf{Filter Pruning}.
In contrast, filter pruning can be well supported by general-purpose hardware and basic linear algebra subprograms (BLAS) libraries, since it removes entire filters without changing the original convolution structures. To this end, Li \emph{et al}.~\cite{li2017pruning} measured filter importance using the weight magnitude. Hu \emph{et al}.~\cite{hu2016network} believed that channels with more sparse outputs are redundant and thus removed the corresponding filters. Lin \emph{et al}.~\cite{lin2020hrank} observed the invariance of feature map rank and removed filters with low-rank feature maps. Molchanov \emph{et al}.~\cite{molchanov2017pruning} adopted Taylor expansion to approximate the influence to the loss function induced by removing each filter. Similarly, \cite{yu2018nisp} optimizes the reconstruction error of the final output response and propagates an ``importance score'' for each channel. \cite{he2017channel} prunes channels using LASSO regression-based selection and the least square reconstruction. Luo \emph{et al}.~\cite{luo2017thinet} established filter pruning as an optimization problem, and removed less important filters based on the statistics of the next layer. In~\cite{liu2017learning}, the scaling factor in the batch normalization (BN) layer is considered as a filter selection indicator to decide whether a filter is important. However, the influence of shifting parameters in the BN layer is totally ignored~\cite{zhao2019variational}. Inspired by this, \cite{kang2020operation} considers both the channel scaling and shifting parameters for pruning. 

\textbf{Discussion}. To the best of our knowledge, only He \emph{et al}.'s work~\cite{he2018soft} implements filter pruning without pretrained models or sparse constraints. However, the main differences between our DCFF and this approach are as below: (1) \cite{he2018soft} picks up ``important'' filters by off-the-shelf $\ell_p$-norm, whereas we propose a fresh Kullback-Leibler divergence-based criterion by exploring the inter-similarity among different filters. (2) \cite{he2018soft} achieves filter pruning in a soft-coded manner where ``unimportant'' filters are zeroized in each forward step; in contrast, our dynamic-coded scheme, as shown in Fig.\,\ref{flow}, does not zeroize any filter, but fuses all filters into a compact set. Also, our DCFF essentially differs from the recent ResRep~\cite{ding2020lossless} that considers convolutional reparameterization for CNN pruning: (1) As stressed across the paper, our DCFF does not introduce any regularization. Differently, ResRep inserts a \textit{compactor} consisting of 1$\times$1 convolution regularized by penalty gradients to indicate filter removal. (2) Our DCFF is to merge all filters into a number of desired ones in the forward propagation while ResRep picks up filters corresponding to non-zero masks. (3) Our filter fusion results in a compact model after training without any post-processing; instead, ResRep requires to reparameterize the introduced \textit{compactor} into the proceeding conv-BN layers after training.

\section{Methodology}\label{methodology}

As discussed in Sec.\,\ref{introduction}, existing filter pruning methods have to pretrain a computation-heavy model, or introduce a hyperparameter-sensitive regularization. In this section, we introduce our DCFF implemented in a computation-economical and regularization-free manner, by detailing its two essential components: dynamic-coded importance and filter fusion, followed by necessary analyses.

\subsection{Preliminary}

Let $M(L^{(1)}, L^{(2)}, ..., L^{(N)})$ be an $N$-layer CNN, where $L^{(i)}$ denotes the $i$-th convolutional layer with a total of $c^{(i)}_{out}$ convolutional filters, which in this paper are represented in a matrix form $\textbf{W}^{(i)} = [\textbf{w}^{(i)}_1, \textbf{w}^{(i)}_2, ..., \textbf{w}^{(i)}_{c^{(i)}_{out}}] \in \mathbb{R}^{d^{(i)} \times c_{out}^{(i)}}$ with $d^{(i)} = c_{in}^{(i)} \cdot w^{(i)} \cdot h^{(i)}$, where $c^{(i)}_{in}$ is the number of input channels, and $w^{(i)}$ and $h^{(i)}$ are the width and height of the filters, respectively. Then, we append the biases of the filters to $\mathbf{W}^{(i)}$, to form a matrix of dimensions $(d^{(i)} + 1) \times c_{out}^{(i)}$. Given its input $\mathbf{O}^{(i-1)}$, \emph{i.e.}, output from the last layer, the output $\mathbf{O}^{(i)}$ of $L^{(i)}$ is computed by:
\begin{equation}\label{origin_output}
    \mathbf{o}^{(i)}_k = \mathbf{w}_k^{(i)} \circledast \mathbf{O}^{(i-1)}, \;\; k = 1, 2, ..., c^{(i)}_{out},
\end{equation}
where $\mathbf{o}^{(i)}_k$ is the $k$-th channel of $\mathbf{O}^{(i)}$ and $\circledast$ denotes the standard convolution operation.

The goal of filter pruning is to derive an $N$-layer compact CNN $\tilde{M}(\tilde{L}^{(1)}, \tilde{L}^{(2)}, ..., \tilde{L}^{(N)})$ with a total of $\tilde{c}^{(i)}_{out}$ filters $\tilde{\textbf{W}}^{(i)} = [\tilde{\textbf{w}}^{(i)}_1, \tilde{\textbf{w}}^{(i)}_2, ..., \tilde{\textbf{w}}^{(i)}_{\tilde{c}^{(i)}_{out}}] \in \mathbb{R}^{\tilde{d}^{(i)} \times \tilde{c}_{out}^{(i)}}$ in $\tilde{L}^{(i)}$ and ideally it should be satisfied that $\tilde{c}^{(i)}_{out} \le c^{(i)}_{out}$. For ease of the representation, the superscript ``$(i)$'' may be dropped from time to time in the following sections.

\begin{figure}[!t]
\begin{center}
\includegraphics[height=0.4\linewidth]{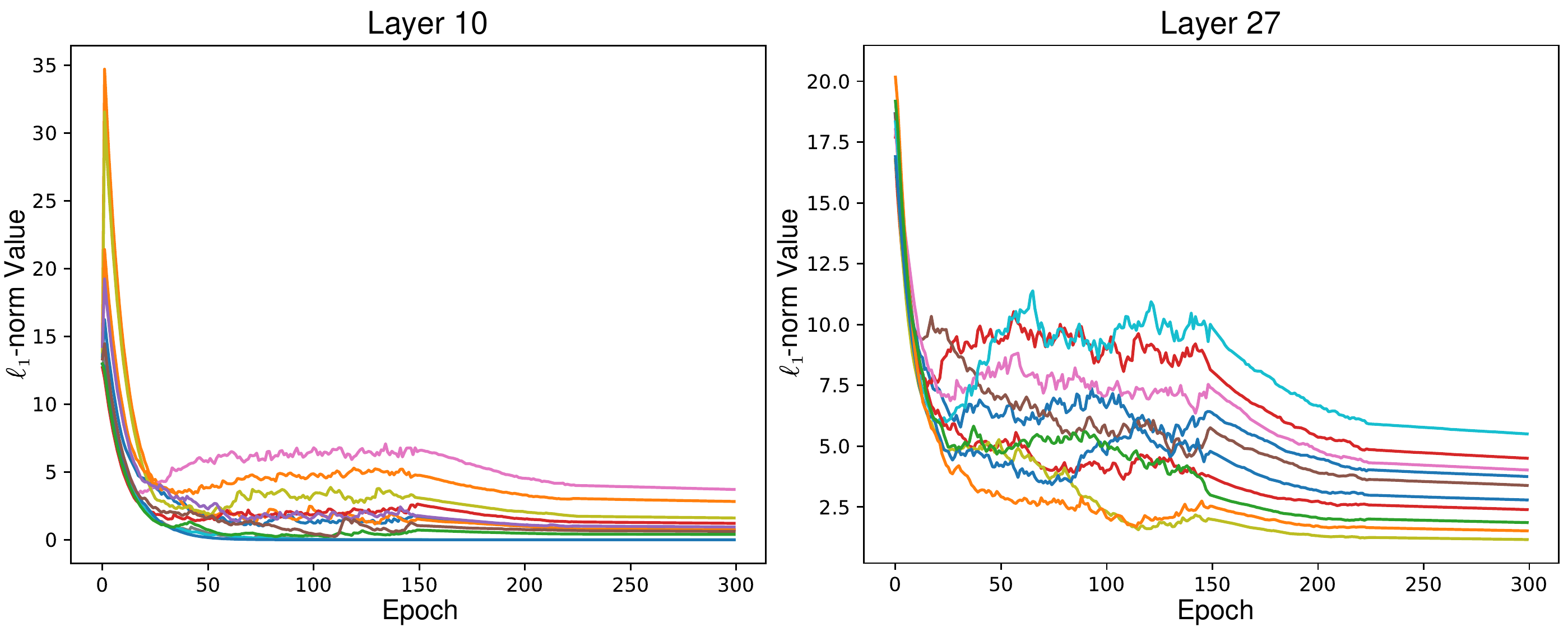}
\end{center}
\caption{\label{magnitude}
The $\ell_1$-norm~\cite{li2017pruning} for different filters (denoted by different colors) at different training epochs (ResNet-56).
}
\end{figure}

\subsection{Dynamic-Coded Importance}\label{dynamic-coded}

Conventional hard-coded methods resort to selecting fixed ``important'' filters upon a pretrained model. We argue that these designs are paradoxical since filters that lead to high performance after fine-tuning no longer follow the high-score standards, as illustrated in Fig.\,\ref{comparison}. The main reasons include two aspects: 1) Although these criteria are indeed the intrinsic property of each filter, the inter-similarity among different filters cannot be well reflected. 2) These criteria are proposed on the basis of a pretrained model. However, as observed in Fig.\,\ref{magnitude}, at different training stages, the relative importance of many filters significantly changes a lot. Besides, after training, the scores among different filters are almost the same (Layer 10). Thus, it is inappropriate to evaluate filter importance based on a pretrained model.

A suitable scenario for measuring filter importance should be constructed on the premise that it can reflect inter-similarity among filters. Also, this scenario should be conducted in a dynamic-coded manner to track real-time importance of each filter during the training of the CNN. Thus, we propose to maintain a distribution $\mathbf{p}_k = (p_{k1}, p_{k2}, ..., p_{kc_{out}})$ as a proxy of $\mathbf{w}_k$. With all probabilities summed up to 1, \emph{i.e.}, $\sum_{j=1}^{c_{out}}p_{kj} = 1$, we define $p_{kj}$ as follows:
\begin{equation}\label{distribution}
    p_{kj} = \frac{\exp({-\mathcal{D}(\mathbf{w}_k, \mathbf{w}_j) \cdot t})}{\sum_{g=1}^{c_{out}} \exp({-\mathcal{D}(\mathbf{w}_k, \mathbf{w}_g) \cdot t})}, \;\; k, j = 1, 2, ..., c_{out},
\end{equation}
where $\mathcal{D}(\cdot, \cdot)$ denotes a distance measurement, and $t$ is a temperature parameter that controls the smoothness of the distribution proxy. While others can be adopted, we find Euclidean distance performs the best as verified in Sec.\,\ref{ablation}.

In particular, the proxy, $\mathbf{p}_k$, standardizes all the distances by transforming each into a probability depending on all the filters, which thus models the inter-similarity between a filter $\mathbf{w}_k$ and other filters in a distribution space. Then, we build the importance of filter $\mathbf{w}_k$ on top of the proxy $\mathbf{p}_k$, instead of the intrinsic property of $\mathbf{w}_k$ such as $\ell_1$-norm~\cite{li2017pruning} or the rank of feature map~\cite{lin2020hrank}. Thus, a natural measurement for $\mathbf{w}_k$ can be defined through the distribution difference between $\mathbf{w}_k$ and others using the Kullback-Leibler (KL) divergence, as defined in the following:
\begin{equation}\label{kl}
    I_{k} = \frac{1}{c_{out}}\sum_{g=1}^{c_{out}}\sum_{j=1}^{c_{out}} p_{kj} \cdot \log\frac{p_{kj}}{p_{gj}}, \;\; k = 1, 2, ..., c_{out}.
\end{equation}

According to the definition of KL-divergence, it is conventional to derive whether the distribution $\mathbf{p}_k$ is different from others. If so, Eq.\,(\ref{kl}) returns a high importance score $I_k$, denoting that $\mathbf{w}_k$ is more important. The rationale lies in that if one filter differentiates a lot from others, it should be representative; otherwise, $\mathbf{w}_k$ can be replaced with its similar counterparts and thus it is less representative. So far, we have derived our inter-similarity standard for selecting $\tilde{c}_{out}$ filters in $\mathbf{W}$ with the highest importance scores.

Then, to realize dynamic-coded importance evaluation along with network training, one naive solution is to re-compute the filter importance before each training epoch, so as to update $\tilde{\mathbf{W}}$. However, this strategy damages the performance as experimentally verified in Sec.\,\ref{analysis}. Specifically, in the early training stage, all filters are initialized randomly and thus they should be authorized equally to compete for important filters. In this case, $\tilde{\mathbf{W}}$ is allowed to be updated drastically. However, the over-frequent updating of the important set $\tilde{\mathbf{W}}$ in the late training stages could unstabilize the network training. Therefore, the relative importance of all filters should be gradually stable as the training continues. To this end, we must adjust the temperature parameter $t$ by formulating it in a training-adaptive manner. Thus, we derive the following:
\begin{equation}\label{temperature}
  t = (T_e - T_s) \cdot \frac{1 + \exp(-E)}{1 - \exp(-E)} \cdot \frac{1 - \exp(-e)}{1 + \exp(-e)} + T_s,
\end{equation}
where $T_s = 1$, $T_e = +\infty$\footnote{$T_e = 10^4$ in our practical implementation.}, $E$ is the total number of training epochs and $e \in [0, E)$ is the current training epoch.

Eq.\,(\ref{temperature}) indicates that, starting with a small value of temperature parameter $t = T_s$ at the beginning of training, the proxy of the distribution $\mathbf{p}_k$ defined in Eq.\,(\ref{distribution}) becomes a soft vector and thus the important score for each filter using Eq.\,(\ref{kl}) can be easily changed, leading to a frequent updating of $\tilde{\mathbf{W}}$. While with an infinite temperature parameter $t = T_e$, $\mathbf{p}_k$ is close to a one-hot distribution vector, where the relative importance score would be gradually stabilized, result of which freezes the updating of $\tilde{\mathbf{W}}$ and stabilizes the training of the network in the end.

\subsection{Filter Fusion}\label{filter_fusion}

By using our dynamic-coded importance described in Sec.\,\ref{dynamic-coded}, we train the compact CNN from scratch to remove the dependency on pretraining a computation-heavy model. In the literature~\cite{li2017pruning,dubey2018coreset,hu2016network,lin2020hrank,molchanov2017pruning,luo2020neural}, a compact filter set $\tilde{\textbf{W}} = [\tilde{\textbf{w}}_1, \tilde{\textbf{w}}_2, ..., \tilde{\textbf{w}}_{\tilde{c}_{out}}]$ is obtained by selecting $\tilde{c}_{out}$ filters with the highest importance scores in $\tilde{\textbf{W}}$ as discussed in Sec.\,\ref{dynamic-coded}, which can be formulated as:
\begin{equation}\label{importance}
\begin{split}
    \tilde{\mathbf{w}}_k = \mathbf{w}_{f(k)}, \;\; k = 1, 2, ..., \tilde{c}_{out},
\end{split}
\end{equation}
where $f(k)$ returns the index $i \in \{1, 2, ..., c_{out}\}$ of the $i$-th filter whose importance score ranks in the $k$-th position.

However, existing methods simply discard low-score filters to obtain the compact filter set, $\tilde{\mathbf{W}}$, and ask for a fine-tuning process in order to pull back the performance. Such a way is even more time-consuming than the cost on the pretrained model when conducted in a layer-wise manner~\cite{he2017channel,luo2017thinet,lin2020hrank}. We believe that despite their low scores, the information of these filters is also crucial to the network performance, since the removal of them leads to significant performance degradation. The fact that filters with large importance values may have small values after fine-tuning (Fig.\,\ref{comparison}) also supports our claim. Thus, a reasonable manner should be that $\tilde{\mathbf{w}}_k$ fuses all information from the original filter set, $\mathbf{W}$, but considers more information from the important filter $\mathbf{w}_{f(k)}$ and less from others rather than directly discarding them. This inspires us to turn back to explore the distribution proxy $\mathbf{p}_{f(k)}$ since it is centered on $\mathbf{w}_{f(k)}$. Under this framework, we can refine the compact filters in Eq.\,(\ref{importance}) as:
\begin{equation}\label{fusion_filter}
\begin{split}
    \tilde{\mathbf{w}}_k = \mathbf{W} \mathbf{p}_{f(k)},  \;\; k = 1, 2, ..., \tilde{c}_{out}.
\end{split}
\end{equation}

\begin{algorithm}[!t]
\caption{\label{alg1}Dynamic-Coded Filter Fusion}
\LinesNumbered
\KwIn{An $N$-layer CNN $M(L^{(1)}, L^{(2)}, ..., L^{(N)})$ with filter sets $\{ \mathbf{W}^{(i)} \}_{i=1}^N$, the number of training epochs $E$, and the number of preserved filter in each layer $\{\tilde{c}_{out}^{(i)}\}_{i=1}^N$.}
\KwOut{A compact CNN $\tilde{M}(\tilde{L}^{(1)}, \tilde{L}^{(2)}, ..., \tilde{L}^{(N)})$ with filter sets $\{ \tilde{\mathbf{W}}^{(i)} \}_{i=1}^N$ and $\tilde{\mathbf{W}}^{(i)} \in \mathbb{R}^{\tilde{d}^{(i)} \times \tilde{c}_{out}^{(i)}}$.}
\For{$e = 0 \rightarrow E$}
{
   Compute the temperature $t$ via Eq.\,(\ref{temperature})\;
   \For{$i = 1 \rightarrow N$}
   {
    \For{$k = 1 \rightarrow c^{(i)}_{out}$}
    {
      Compute the distribution proxy $\mathbf{p}_k$ for filter $\mathbf{w}^{(i)}_{k}$ via Eq.\,(\ref{distribution});
    }
    \For{$k = 1 \rightarrow c^{(i)}_{out}$}
    {
      Get the importance score for filter $\mathbf{w}^{(i)}_{k}$ via Eq.\,(\ref{importance})\;
    }
    Compute the fused filter set $\tilde{\mathbf{W}}^{(i)}$ via Eq.\,(\ref{fusion_filter})\;
   }
   Forward the input image batch using the fused filter set $\{ \tilde{\mathbf{W}}^{(i)} \}_{i=1}^N$ via Eq.\,(\ref{compact_convolution})\;
   Update the original filter set  $\{ \mathbf{W}^{(i)} \}_{i=1}^N$\;
}
\end{algorithm}

Therefore, each fused filter, $\tilde{\mathbf{w}}_k$, is a linear combination of all filters in $\mathbf{W}$, \emph{i.e.}, the weighted average regarding the distribution $\mathbf{p}_{f(k)}$. The innovation of our filter fusion can be explained via the training-adaptive temperature parameter. Specifically, a small temperature smooths the proxy $\mathbf{p}_{f(k)}$, which thus integrates more information from all filters in $\mathbf{W}$. As the training proceeds, $\mathbf{p}_{f(k)}$ gradually approximates to a one-hot vector centered on $\mathbf{w}_{f(k)}$, and then our fusion formulation in Eq.\,(\ref{fusion_filter}) becomes Eq.\,(\ref{importance}). It can be seen that our filter fusion is a generalization of Eq.\,(\ref{importance}).

In the forward step, we first update the temperature parameter so as to re-compute the compact filter set $\tilde{\mathbf{W}}$. Then,  the convolution in the $i$-th layer (Eq.\,(\ref{origin_output})) under our compact training framework can be reformulated as:
\begin{equation}\label{compact_convolution}
\begin{split}
    \tilde{\mathbf{o}}^{(i)}_k &= \tilde{\mathbf{w}}_k^{(i)} \circledast \tilde{\mathbf{O}}^{(i-1)}
    \\&
    =\big(\mathbf{W}^{(i)}\mathbf{p}_{f(k)}\big) \circledast \tilde{\mathbf{O}}^{(i-1)}, \;\; k = 1, 2, ..., \tilde{c}^{(i)}_{out}.
\end{split}
\end{equation}

As shown in Fig.\,\ref{flow}, for the backpropagation, we update the original filters $\mathbf{W}^{(i)}$ via the chain rule. After a standard network training without any sparse constraint, the compact filter sets for all layers $\{ \mathbf{W}^{(i)} \}_{i=1}^N$ are then preserved for inference, which greatly facilitates the practical deployment of filter pruning  and differentiates our DCFF from existing regularized-retraining studies.

We summarize the main steps of our dynamic-coded filter fusion for training compact CNNs in Alg.\,\ref{alg1}.

\begin{table*}[!t]
\caption{\label{cifar}Quantitative results on CIFAR-10. We report the top-1 classification accuracy, the FLOPs, the amount of parameters, and the pruning rate of the compact models.}
\centering
\begin{tabular}{c|cc|cc|cc}
\toprule
Method    &Top1-acc &$\uparrow\downarrow$  &FLOPs  &Pruning Rate  &Parameters &Pruning Rate\\ \hline
VGGNet-16~\cite{simonyan2015very}     &93.02\%    &0.00\%    &314.59M &0.0\% &14.73M     &0.0\%    \\
SSS~\cite{huang2018data}           &93.02\%    &0.00\%          &183.13M &41.6\% &3.93M      &73.8\%    \\ 
Zhao \emph{et al}.~\cite{zhao2019variational} &93.18\%    &0.16\%$\uparrow$        &190.00M &39.1\% &3.92M      &73.3\%    \\
HRank~\cite{lin2020hrank}   &92.34\%    &0.38\%$\downarrow$  &108.61M &65.3\% &2.64M      &82.1\%    \\
Hinge~\cite{li2020group} &92.91\%&0.11\%$\downarrow$&191.68M&39.1\%&2.94M&80.1\% \\
SWP~\cite{meng2020pruning} &92.85\%&0.17\%$\downarrow$&90.60M&71.2\%&1.08M&92.7\% \\
\textbf{DCFF} (Ours)
&\textbf{93.47\%}&\textbf{0.45\%$\uparrow$}&\textbf{72.77M}&\textbf{76.8\%}&\textbf{1.06M}&\textbf{92.8\%} \\
\hline
GoogLeNet~\cite{szegedy2015going} &95.05\%  &0.00\%   &1.53B &0.00\% &6.17M   &0.00\%    \\
L1~\cite{li2017pruning}    &94.54\%  &0.51\%$\downarrow$      &1.02B    &32.9\% &3.51M   &42.9\%    \\
ABCPruner~\cite{lin2020channel} &94.84\%  &0.21\%$\downarrow$ &0.51B &66.6\% &2.46M &60.1\% \\
HRank~\cite{lin2020hrank} &94.53\%  &0.52\%$\downarrow$      &0.49B    &67.9\% &2.18M   &64.7\%    \\      
\textbf{DCFF} (Ours) &\textbf{94.92\%}&\textbf{0.13\%}$\downarrow$&\textbf{0.46B}&\textbf{70.1\%}&\textbf{2.08M}&\textbf{66.3\%} \\
\hline
ResNet-56~\cite{he2016deep}  &93.26\%    &0.00\%    &127.62M  &0.0\% &0.85M    &0.0\%       \\
L1~\cite{li2017pruning}   &93.06\%    &0.20\%$\downarrow$      &90.90M &27.6\%  &0.73M  &14.1\%    \\
NISP~\cite{yu2018nisp} &93.01\%    &0.25\%$\downarrow$       &81.00M &35.5\%  &0.49M  &42.4\%    \\
FPGM~\cite{he2019filter} &93.26\% &0.00\%$\downarrow$ &59.40M &52.6\% &- &- \\
LFPC~\cite{he2020learning} &93.24\% &0.02\%$\downarrow$ &59.10M &52.9\% &- &- \\
HRank~\cite{lin2020hrank} &93.17\%    &0.09\%$\downarrow$    &62.72M &50.0\%  &0.49M 
&42.4\%   \\
SCP~\cite{kang2020operation} &93.23\% &0.03\%$\downarrow$ &61.89M &51.5\% &0.44M &48.4\%  \\
\textbf{DCFF} (Ours) &\textbf{93.26\%}&\textbf{0.00\%}&\textbf{55.84M}&\textbf{55.9\%}&\textbf{0.38M}&\textbf{55.0\%} \\
\hline
ResNet-110~\cite{he2016deep} &93.50\%    &0.00\%    &257.09M  &0.0\% &1.73M   &0.0\%    \\
L1~\cite{li2017pruning} &93.30\%    &0.20\%$\downarrow$         &155.00M  &38.7\% &1.16M   &32.6\%    \\
HRank~\cite{lin2020hrank} &93.36\%    &0.14\%$\downarrow$        &105.70M  &58.2\% &0.70M   &59.2\%    \\
LFPC~\cite{he2020learning} &93.07\% &0.43\%$\downarrow$ &101.00M &60.3\% & - &- \\
\textbf{DCFF} (Ours) &\textbf{93.80\%}&\textbf{0.30\%$\uparrow$}&\textbf{85.30M}&\textbf{66.6\%}&\textbf{0.56M}&\textbf{67.9\%} \\
\bottomrule
\end{tabular}
\end{table*}

\section{Experiments}\label{experiment}
%
%

To demonstrate the ability of the proposed DCFF, in this section, we conduct model pruning for representative networks, including VGGNet-16~\cite{simonyan2015very}, GoogLeNet~\cite{szegedy2015going} and ResNet-56/110~\cite{he2016deep} on CIFAR-10~\cite{krizhevsky2009learning}. In addition, we train compact versions of ResNet-50~\cite{he2016deep} and MobileNet-V2~\cite{sandler2018mobilenetv2} on the large-scale ILSVRC-2012~\cite{russakovsky2015imagenet}. 
We manually determine the pruned filter number $\tilde{c}^{(i)}_{out}$ in this paper, and to ensure the reproducibility, we have provided all per-layer pruning ratios in our project link at \url{https://github.com/lmbxmu/DCFF}. 
Note that our method is complementary to the recent ABCPruner~\cite{lin2020channel} and EagleEye~\cite{li2020eagleeye} that adopt structure searching or global ranking to find a per-layer pruning ratio. Therefore, these methods can be considered as a supplementary means for determining per-layer pruning ratios without the necessity of human involvement. Usually, better performance can be observed as demonstrated in Sec.\,\ref{ablation}.

\subsection{Training Settings}\label{settings}
We train our compact CNN models from scratch using the SGD optimizer with a momentum of 0.9 and the batch size is set to 256. On CIFAR-10, we train the compact CNNs for a total of 300 epochs and the weight decay is set to 5$\times$10$^{\text{-4}}$. The learning rate is initially set to 0.1, and then divided by 10 at the training points of 150 and 225 epochs. 
On ILSVRC-2012, a total of 90 epochs are given to train compact ResNet-50 with the weight decay set to 1$\times$10$^\text{-4}$, and the initial learning rate is set to 0.1, which is then multiplied by 0.1 at the points of 30 and 60 training epochs. 
Besides, following~\cite{ding2020lossless,luo2020autopruner,luo2020neural}, we also consider the cosine scheduler~\cite{Loshchilov2017SGDRSG} to adjust the learning rate for ResNet-50/MobileNet-V2~\cite{sandler2018mobilenetv2} with the weight decay set to 1$\times$10$^\text{-4}$/4$\times$10$^\text{-5}$. The initial learning rate is set to 1$\times$10$^\text{-2}$/1$\times$10$^\text{-1}$ for ResNet-50/MobileNet-V2~\cite{sandler2018mobilenetv2}. Also, the training epochs for MobileNet-V2 are 180.

For fair comparison, all methods are fed with random crops and horizontal flips of the training images, which are also official operations in Pytorch\footnote{More details can be referred to \url{https://github.com/pytorch/examples/blob/master/imagenet/main.py}.}. Nevertheless, other data augmentation techniques such as lightening and color jitter in the source code of~\cite{yu2019slimmable,liu2019metapruning,li2020eagleeye} can be applied to further boost the pruned model performance.

\subsection{Performance Metrics}\label{metric}

For quantitative comparison, we report four widely-used metrics including accuracy, FLOPs, parameters, and pruning rate. Following most off-the-shelf pruning methods, for CIFAR-10, we report the top-1 accuracy of the pruned models. As for ILSVRC-2012, we report both top-1 and top-5 classification accuracies.

\subsection{Results on CIFAR-10}\label{results_cifar}

\textbf{VGGNet}~\cite{simonyan2015very}. We start by applying our DCFF to train a compact VGGNet-16. As displayed in Table\,\ref{cifar}, our DCFF achieves 93.47\% top-1 accuracy meanwhile removing 76.8\% FLOPs and 92.8\% parameters. DCFF significantly outperforms its existing competitors and leads to a large reduction of the model complexity.

\textbf{GoogLeNet}~\cite{szegedy2015going}. In Table\,\ref{cifar}, compared to the state-of-the-art HRank~\cite{lin2020hrank}, our DCFF shows its capacity to maintain a higher accuracy (94.92\%~\emph{vs}.~94.53\%) meanwhile reducing more FLOPs (70.1\%~\emph{vs}.~67.9\%) and parameters (66.3\%~\emph{vs}.~64.7\%). It is worth noting that HRank heavily relies on expensive model pretraining and fine-tuning. In contrast, our DCFF simply trains a compact model from scratch, resulting in a major advantage of reducing great number of processing time.

\textbf{ResNet-56/110}~\cite{he2016deep}. We continue to train compact ResNets using different depths of 56 and 110. We can see from Table\,\ref{cifar} that, with more reductions of both FLOPs and parameters, the proposed DCFF well retains the performance of the original ResNet-56 and further increases the accuracy of ResNet-110 by 0.30\%. These results are significantly better than off-the-shelf counterparts.

\subsection{Results on ILSVRC-2012}\label{results_ilsvrc}
%

We also conduct experiments on the large-scale ILSVRC-2012 for training compact ResNet-50~\cite{he2016deep} in Table\,\ref{res50}. For fair comparison, we perform our DCFF with different pruning rates such that the accuracy can be compared under a similar complexity reduction.

\textbf{ResNet-50}~\cite{he2016deep}. The compared SOTAs for ResNet-50 in Table\,\ref{res50} are HRank~\cite{lin2020hrank}, LFPC~\cite{he2020learning}, ResRep~\cite{ding2020lossless}, AutoPruner~\cite{luo2020autopruner} and CURL~\cite{luo2020neural}. Compared with them, our DCFF achieves higher test accuracy while more FLOPs and parameters are reduced. For example, our DCFF achieves 75.18\% top-1 and 92.56\% top-5 accuracies after pruning 45.3\% FLOPs and removing 40.7\% parameters, which are better than ABCPruner that retains the accuracies of 74.84\% and 92.31\% on top-1 and top-5 after reducing 40.8\% FLOPs and 33.8\% parameters. In comparison with CURL that obtains 73.39\% top-1 and 91.46\% top-5 accuracies with the reductions of 73.2\% FLOPs and 73.9\% parameters, our DCFF retains better top-1 accuracy of 73.81\% and top-5 accuracy of 91.59\%, and meanwhile, it reduces more FLOPs of 75.1\% and more parameters of 74.3\%. These results verify the effectiveness of our dynamic-coded filter fusion in training a compact CNN model even on a large-scale dataset.

\begin{table*}[!t]
\caption{\label{res50}Quantitative results on ILSVRC-2012. We report the top-1 and top-5 accuracy, the FLOPs, the amount of parameters, and the pruning rate of the compact models. $^{\ast}$ shows the learning rate with the cosine scheduler. DCP$^*$ denotes our reproduced results by removing the reconstruction error.}
\centering
\begin{tabular}{c|cc|cc|cc|cc}
\toprule
Method   &Top1-acc &$\uparrow\downarrow$ &Top5-acc &$\uparrow\downarrow$ &FLOPs  &Pruning Rate  &Parameters &Pruning Rate \\ \hline
ResNet-50~\cite{he2016deep} &76.15\%&0.00\%&92.96\%&0.00\%&4.14B&0.0\%&25.56M&0.0\% \\
ThiNet-30~\cite{luo2017thinet} &68.42\% &7.59\%$\downarrow$  &88.30\%&4.66\%$\downarrow$  &1.10B&73.4\%      &8.66M &66.1\% \\
HRank~\cite{lin2020hrank} &69.10\%&6.91\%$\downarrow$  &89.58\%&3.38\%$\downarrow$ &0.98B   &76.3\%     &8.27M &67.6\%\\
\textbf{DCFF} (Ours) &\textbf{71.54\%}&\textbf{4.53\%$\downarrow$}&\textbf{90.57\%}&\textbf{2.39\%$\downarrow$} &\textbf{0.96B} &\textbf{76.7\%}  &\textbf{7.40M}  &\textbf{71.0\%} \\
SSS-26~\cite{huang2018data} &71.82\%&4.19\%$\downarrow$  &90.79\% &2.17\%$\downarrow$ &2.33B &43.7\%    &15.60M &39.0\%\\
HRank~\cite{lin2020hrank} &71.98\%&4.17\%$\downarrow$  &91.01\% &1.95\%$\downarrow$ &1.55B &62.6\%  &13.37M &47.7\%  \\
ABCPruner~\cite{lin2020channel}&73.52\%&2.63\%$\downarrow$&91.51\%&1.45\%$\downarrow$&1.79B&56.6\%&11.24M&56.0\%\\
LFPC~\cite{he2020learning} &74.18\%&1.97\%$\downarrow$&91.92\%&1.04\%$\downarrow$&1.60B&61.4\%&-&- \\
NPPM~\cite{gao2021network} &75.96\%&0.19\%$\downarrow$&92.75\%&0.21\%$\downarrow$&2.32B&56.0\%&-&- \\
DCP~\cite{liu2021discrimination} &74.99\%&1.16\%$\downarrow$&92.20\%&0.76\%$\downarrow$&2.13B&51.5\%&14.19M&55.5\% \\
DCP$^*$~\cite{liu2021discrimination} &74.13\%&2.02\%$\downarrow$&91.87\%&1.09\%$\downarrow$&2.13B&51.5\%&14.19M&55.5\% \\
\textbf{DCFF} (Ours) &\textbf{74.21\%}&\textbf{1.94\%$\downarrow$}&\textbf{91.93\%}&\textbf{1.03\%$\downarrow$} &\textbf{1.49B}&\textbf{63.8\%}&\textbf{10.58M}&\textbf{58.6\%} \\
SSS-32~\cite{huang2018data}&74.18\%&1.97\%$\downarrow$  &91.91\% &1.05\%$\downarrow$  &2.82B   &31.9\%   &18.60M &27.2\% \\
CP~\cite{he2017channel}&72.30\%&3.85\%$\downarrow$ &90.80\%  &2.16\%$\downarrow$  &2.73B &34.1\%   &-&-\\
SFP~\cite{he2018soft} &74.61\%&1.54\%$\downarrow$&92.06\%&0.90\%$\downarrow$&2.41B&41.8\%&-&- \\
ABCPruner~\cite{lin2020channel}&74.84\%&1.31\%$\downarrow$&92.31\%&0.65\%$\downarrow$&2.45B&40.8\%&16.92M&33.8\% \\
\textbf{DCFF} (Ours) &\textbf{75.18\%} &\textbf{0.97\%$\downarrow$} &\textbf{92.56\%}  &\textbf{0.50\%$\downarrow$} &\textbf{2.25B}&\textbf{45.3\%} &\textbf{15.16M} &\textbf{40.7\%}\\
ResRep$^{\ast}$~\cite{ding2020lossless} &75.49\%&0.66\%$\downarrow$&92.55\%&0.32\%$\downarrow$&1.55B&62.1\%&-&- \\   
\textbf{DCFF$^{\ast}$} (Ours) &\textbf{75.60\%}&\textbf{0.55\%$\downarrow$}&\textbf{92.55\%}&\textbf{0.32\%$\downarrow$}&\textbf{1.52B}&\textbf{63.0\%}&\textbf{11.05M}&\textbf{56.8\%} \\
AutoPruner$^{\ast}$~\cite{luo2020autopruner} &73.05\%&3.10\%$\downarrow$  &91.25\%  &1.71\%$\downarrow$   &1.39B   &66.4\%   &12.69M &50.4\%\\
\textbf{DCFF$^{\ast}$} (Ours) &\textbf{74.85\%}&\textbf{1.30\%}$\downarrow$&\textbf{92.41\%}&\textbf{0.55\%$\downarrow$}&\textbf{1.38B}&\textbf{66.7\%}&\textbf{11.81M}&\textbf{53.8\%} \\
CURL$^{\ast}$~\cite{luo2020neural}&73.39\%&2.76\%$\downarrow$&91.46\%&1.50\%$\downarrow$&1.11B&73.2\%&6.67M&73.9\%\\
\textbf{DCFF$^{\ast}$} (Ours) &\textbf{73.81\%}&\textbf{2.34\%$\downarrow$} &\textbf{91.59\%} &\textbf{1.37\%$\downarrow$} &\textbf{1.02B}&\textbf{75.1\%}&\textbf{6.56M}&\textbf{74.3\%}  \\
\hline
MobileNet-V2~\cite{sandler2018mobilenetv2} &72.00\% &0.00\% &90.12\% &0.00\% &300M  &0.0\% &3.50M &\color{blue}0.0\% \\
MetaPruning~\cite{liu2019metapruning}&68.20\% &3.80\%$\downarrow$&87.96\%&2.16\%$\downarrow$ &140M &53.3\% &2.62M	&25.1\%\\
DCFF (Ours) &\textbf{68.60\%}&\textbf{3.40\%$\downarrow$}&\textbf{88.13}\%&\textbf{1.99}\%$\downarrow$&\textbf{140M} &\textbf{53.3\%} &\textbf{2.62M}	&\textbf{25.1\%}\\
\bottomrule
\end{tabular}
\end{table*}

Note that recent methods including NPPM~\cite{gao2021network} and DCP~\cite{liu2021discrimination} seem to perform better in Table\,\ref{res50}. Here, we would like to stress some factualities. Regarding NPPM, no parameter reduction has been reported. Empirically, more parameters lead to better performance. It is hard to tell if better performance of NPPM comes from more parameters. As for DCP, the performance gap becomes much smaller. Nevertheless, our DCFF reaches more complexity reduction of 58.60\% \emph{vs.} 51.55\% in FLOPs and 63.8\% \emph{vs.} 55.50\% in parameters. Also, we would like to stress that DCP implicitly utilizes knowledge distillation where a recontruction error between the pre-trained model (teacher) and the pruned model (student) is built. For a more fair comparison, we reproduce NPPM based on the released code but removing the recontruction error, denoted as DCP$^*$. Consequently, under the same complexity reduction, we obtain 74.13\% in the top-1 accuracy and 91.87\% in the top-5 accuracy. Also, the pruning cost of DCP is very expensive where a total of 8 days are used to finish our reproduction including 3 days for pre-training a ResNet-50, 3 days for pruning the network layer-wisely and 2 days for fine-tuning the pruned model. In contrast, it takes around 2 days to complete each experiment of our DCFF.

\textbf{MobileNet-V2}~\cite{sandler2018mobilenetv2}. In contrast to ResNet-50, compression on MobileNet is more challenging for its extremely compact design.
For fairness, we inherit the per-layer pruning ratio from MetaPruning~\cite{liu2019metapruning} such that the pruned model resides in the same complexity. We perform our DCFF and manifest the compared results in Table\,\ref{res50}.
We can see our DCFF achieves lower top-1 errors than its counterparts. 
As against MetaPruning that has only 68.20\% accuracy in the top-1, DCFF returns higher performance of 68.60\%. 

\subsection{Performance Analysis}\label{analysis}

\begin{figure}[!t]
\begin{center}
\includegraphics[height=0.47\linewidth]{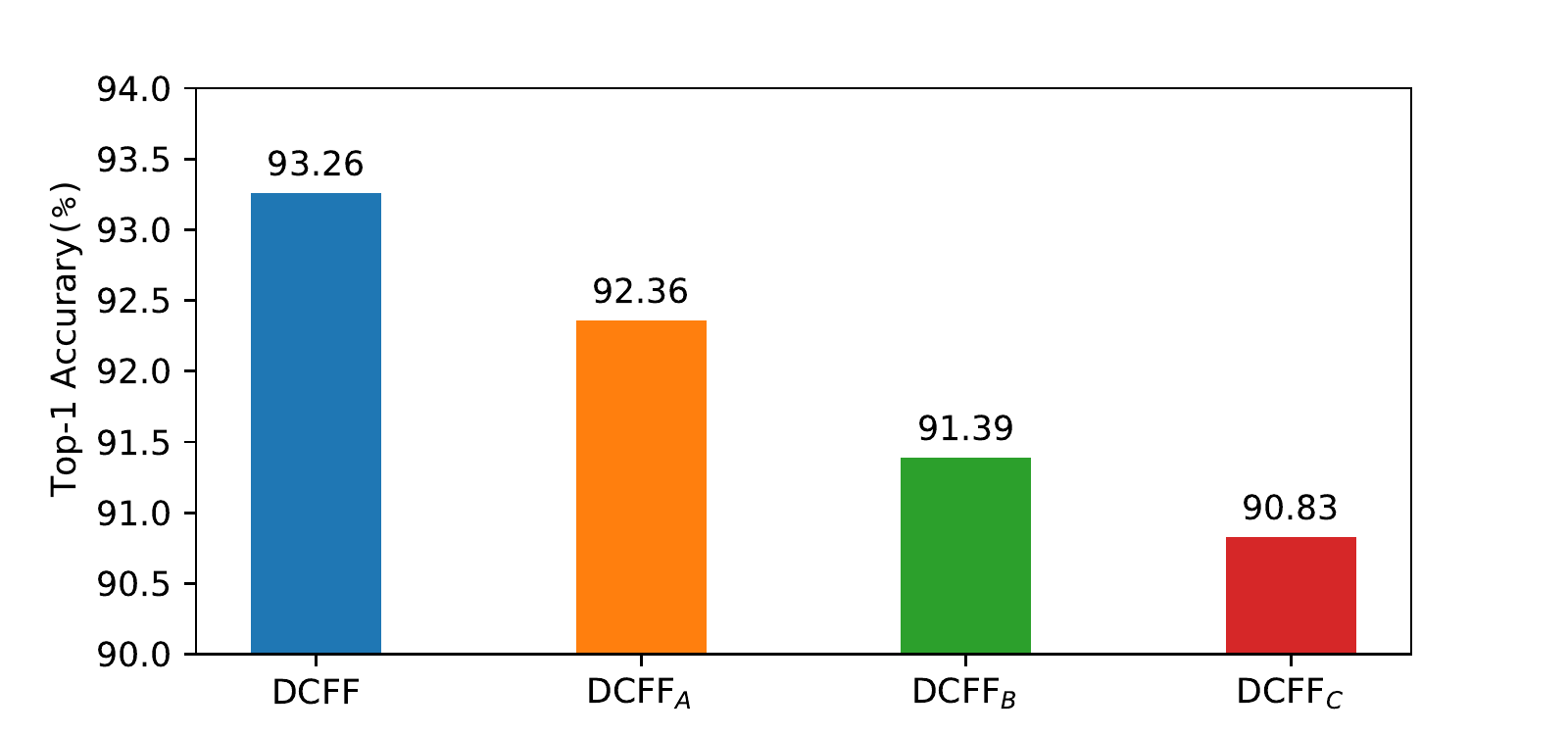}
\caption{\label{variant}
Top-1 accuracy of ResNet-56 for the variants of DCFF on CIFAR-10. DCFF$_A$ uses $\ell_1$-norm as filter importance. DCFF$_B$ has no filter fusion. DCFF$_C$ uses a fixed temperature $t = 1$.
}
\end{center}
\end{figure}

To analyze the proposed method, we develop three variants of DCFF, including: (1) DCFF$_A$: We measure the filter importance using the intrinsic property-based $\ell_1$-norm to replace our inter-similarity-based importance criterion defined in Eq.\,(\ref{kl}). (2) DCFF$_B$: The filter fusion proposed in Sec.\,\ref{filter_fusion} is removed. We simply preserve the high-score filters for training while the low-score filters are discarded. (3) DCFF$_C$: We replace the training-adaptive temperature parameter $t$ in Eq.\,(\ref{temperature}) with a constant $t = 1$. Note that, to pursue fair comparison among all variants, we adopt the same pruning rate for ResNet-56 in Table\,\ref{cifar}, that is, 55.9\% FLOPs are reduced and 55.0\% parameters are removed. Then, we report the top-1 accuracy in Fig.\,\ref{variant}.

Fig.\,\ref{variant} shows that DCFF achieves the best performance, with the top-1 accuracy of 93.26\%. Then, by replacing our KL-divergence-based filter importance, which reflects the inter-similarity among filters with $\ell_1$-norm that essentially measures the intrinsic property of each filter, DCFF$_A$ decreases the performance to 92.36\%, resulting in 0.9\% accuracy drop. It well demonstrates that the inter-similarity-based evaluation can capture the relative importance of filters more accurately. Further, we explore the effect of our filter fusion. As can be observed, without the involvement of the filter fusion (DCFF$_B$), the accuracy decreases to 91.39\%, showing that low-score filters also do benefit to the accuracy performance of compact networks.

\begin{figure*}[th]
\begin{center}
\begin{minipage}[t]{0.48\linewidth}
\centerline{
\subfigure[Training with the adaptive temperature $t$ using Eq.\,(\ref{temperature}).]{
\includegraphics[width=\linewidth]{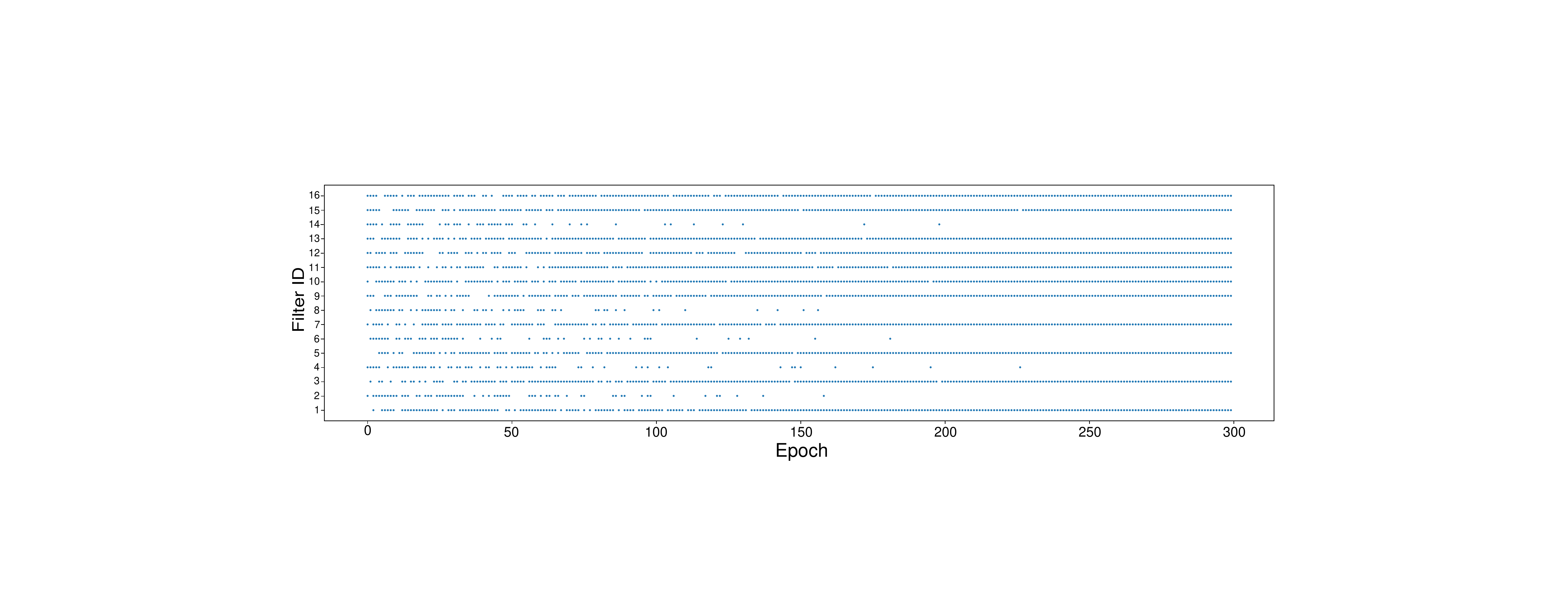}}
\subfigure[Training with the fixed temperature $t = 1$.]{
\includegraphics[width=\linewidth]{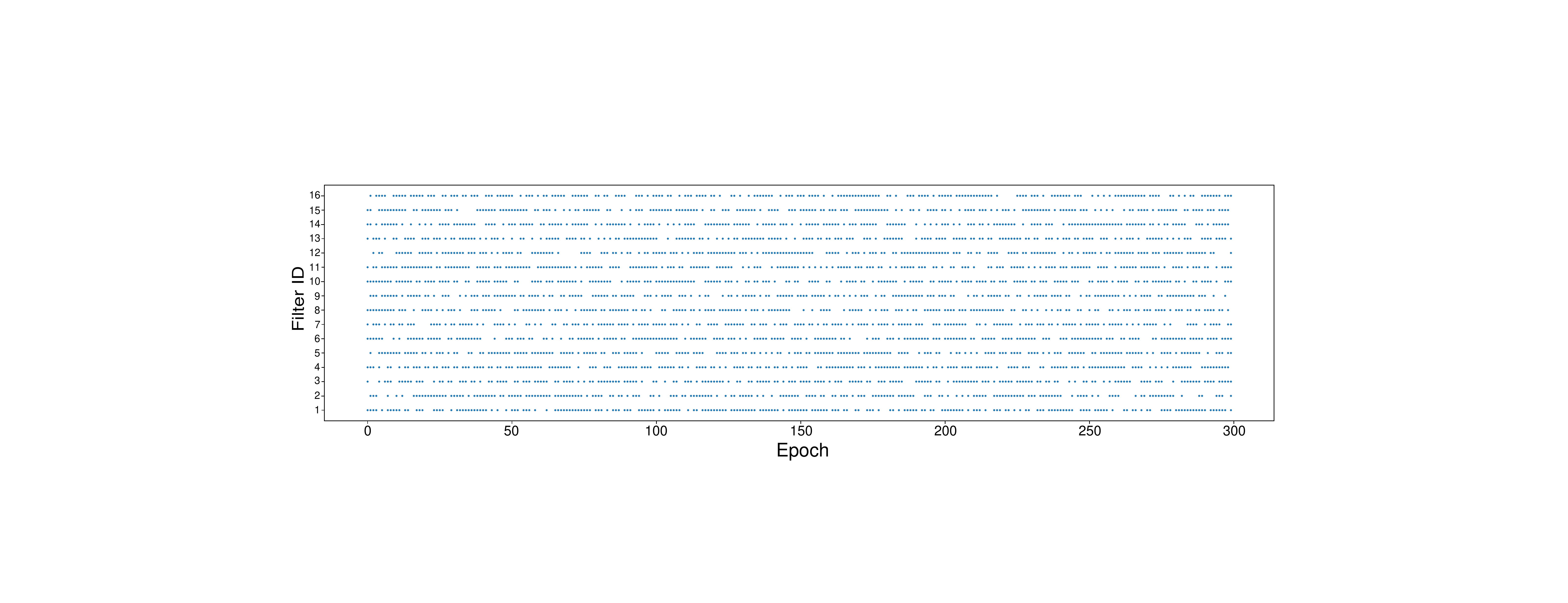}}\hspace*{0.08\linewidth}
}
\end{minipage}
\caption{\label{temperature_analysis}Analysis of the effect of the temperature $t$ with and without our training-adaptive formulation in Eq.\,(\ref{temperature}). The blue dots denote high-score filters in each training epoch. Experiments are conducted using ResNet-56 (Layer 11). A total of 11 filters are preserved.}
\end{center}
\end{figure*}

Lastly, we illustrate the necessity of using the training-adaptive temperature parameter $t$. Setting $t = 1$ (DCFF$_C$) leads to a significant accuracy drop of 2.43\% in comparison with our training-adaptive scheme. To dive into a deeper analysis, in Fig.\,\ref{temperature_analysis}, we visualize the high-score filters in different training epochs. The high-score filters drastically change at the beginning of the network training for both the temperature designs. As the training goes on, with $t = 1$, the high-score filters still retain a drastic change which damages the network performance as discussed in Sec.\,\ref{dynamic-coded}, whilst our training-adaptive formulation gradually fixes the relative importance of filters in the late training stages and thus stabilizes the network training.

\subsection{Ablation Study}\label{ablation}
In this subsection, we continue to provide some ablations including the distance measurement in Eq.\,(\ref{distribution}), filter fusion in Eq.\,(\ref{kl}), the temperature $T_e$ in Eq.\,(\ref{temperature}), as well as per-layer pruning ratio. All experiments are performed upon ILSVRC-2012 using ResNet-50 of 2.25B FLOPs and 15.16M parameters from Table\,\ref{res50}.

Recall in Eq.\,(\ref{distribution}), we measure filter distance using Euclidean distance. In Table\,\ref{distance}, we compare with other distance measurements including Manhattan distance, Correlation distance, and Cosine distance. The results manifest that the Euclidean distance shows the best result among all. The potential reason for the poor performance of compared methods might be attributed to our observation that these distance measurements change more drastically than the commonly-used Euclidean distance, leading to unstable network training.
Therefore, across the paper, the Euclidean distance is adopted to perform all experiments.

\begin{table}[!t]
\centering
\caption{Influence of distance measurement to the performance of pruned ResNet-50. We report the top-1 accuracy on ILSVRC-2012.}
\label{distance}
\begin{tabular}{c|c|c|c}
\toprule
Euclidean & Manhattan & Correlation & Cosine \\ \hline
75.18     & 74.72     & 74.91       & 74.93  \\ 
\toprule
\end{tabular}
\end{table}

\begin{table}[!t]
\centering
\caption{Performance of pruned ResNet-50 using different fusion methods. ``Inversely'' indicates the filter importance as $-I_k$ in Eq.\,(\ref{kl}). ``Randomly'' denotes a random measurement in each forward propagation. We report the top-1 accuracy on ILSVRC-2012.}
\label{fusion}
\begin{tabular}{c|c|c}
\toprule
Ours & Inversely & Randomly \\ \hline
75.18    & 74.73     & 55.63     \\ 
\toprule
\end{tabular}
\end{table}

We continue with the analyses on our filter fusion method. Eq.\,(\ref{kl}) measures the filter importance. For comparison, we introduce two variants: First, we inversely measure the importance of each filter as $-I_k$. Second, the filter importance is randomly measured in each forward propagation. From Table\,\ref{fusion}, we can see that the compared fusion methods lead to performance degradation. In particular, the performance of ``Randomly'' severely drops to 55.63\%, indicating that filter fusion requires a heuristic guidance. As analyzed in Sec.\,\ref{dynamic-coded}, a large $I_k$ indicates $\mathbf{w}_k$ is more representative, which however is removed by ``Reversely', therefore, performance drop occurs as well.

We finally study the influence of temperature parameter $T_e$. According to Eq.\,(\ref{temperature}), $T_e$ is expected to be infinite for the purpose of a one-hot distribution vector. Table\,\ref{te} manifests the performance \emph{w.r.t.} different values of $T_e$ in the experimental implementation. For a small $T_e$, we observe severe performance drops since it destroys one-hot distribution and then unstabilizes the relative importance score. Better accuracy can be obtained by increasing $T_e$ until around $10^4$ which is large enough to maintain a one-hot distribution and therefore the performance becomes stable.

\begin{table}[h]
\centering
\caption{Performance of pruned ResNet-50 \emph{w.r.t.} different values of temperature parameter $T_e$. We report the top-1 accuracy on ILSVRC-2012.}
\label{te}
\begin{tabular}{c|c|c|c|c}
\toprule
$T_e$ & $10^1$ & $10^2$ &$10^4$ &$10^8$ \\ \hline
Accuracy   &69.55    &73.93  &75.18 &75.16    \\ 
\toprule
\end{tabular}
\end{table}

\begin{table}[h]
\centering
\caption{Performance comparison of pruned ResNet-50 between manually defined per-layer pruning ratio and structure searching~\cite{lin2020channel}/global ranking~\cite{li2020eagleeye}. We report the top-1 accuracy on ILSVRC-2012.}
\label{per-layer}
\begin{tabular}{c|c|c|c}
\toprule
$T_e$ & Accuracy & FLOPs & Parameters  \\ \hline
DCFF$_{manually}$   &71.54    &960M  &7.40M\\ 
DCFF$_{ABCPruner}$   &72.19    &945M  &7.35M\\
\toprule
DCFF$_{manually}$   &74.21    &1490M  &10.58M\\ 
DCFF$_{ABCPruner}$   &73.78    &1295M  &9.10M\\
\toprule
DCFF$_{manually}$   &74.21    &1490M  &10.58M\\ 
DCFF$_{ABCPruner}$   &74.73    &1794M  &11.24M\\
\toprule
DCFF$_{manually}$   &75.18    &2250M  &15.16M\\ 
DCFF$_{ABCPruner}$   &74.83    &1891M  &11.75M\\
\toprule
DCFF$_{manually}$   &75.18    &2250M  &11.75M\\ 
DCFF$_{ABCPruner}$   &75.79    &2256M  &18.02M\\
\toprule
DCFF$_{manually}$   &71.54    &960M  &7.40M\\ 
DCFF$_{EagleEye}$   &73.18    &1040M  &6.99M\\
\toprule
DCFF$_{manually}$   &75.18   &2250M  &15.16M\\ 
DCFF$_{EagleEye}$   &75.60    &2070M  &14.41M\\
\toprule
\end{tabular}
\end{table}

As stated in the beginning of Sec.\,\ref{experiment}, we manually determine the per-layer pruning ratios. Nevertheless, recent structure searching based ABCPruner~\cite{lin2020channel} and global ranking based EagleEye~\cite{li2020eagleeye} can be a supplementary means for determining per-layer pruning ratios without human involvement. To verify this, we borrow the per-layer pruning ratios from the released checkpoints of ResNet-50~\cite{lin2020channel,li2020eagleeye} and perform our DCFF. Table\,\ref{per-layer} provides performance comparison between our manually defined per-layer pruning ratios and these based on structure searching~\cite{lin2020channel}/global ranking~\cite{li2020eagleeye}.  As can be seen, structure searching and global ranking mostly return better performance under similar model complexity reduction. Therefore, the proposed method can be effectively boosted by appropriately combining these studies that are devoted to better per-layer pruning ratios.

\section{Conclusion}\label{conclusion}
In this paper, a novel dynamic-coded filter fusion (DCFF) is introduced to train compact CNNs. The method successfully realizes the CNN pruning without the dependency on a computation-heavy pretrained model and the introduction of hyperparameter-sensitive sparsity constraints. To this end, we first maintain a distribution as a proxy of each filter, on top of which, an inter-similarity importance evaluation is devised to measure the relative importance of filters. The distribution proxy gradually approximates to a one-hot vector as its temperature parameter approaches infinity, leading to a dynamic-coded importance evaluation. Furthermore, instead of simply abandoning low-score filters, we propose to fuse all filters using the assigned distribution proxy as our preserved filters in the forward propagation. In the backward, the original filters are updated by the SGD optimizer. After a simple network training from scratch, we preserve the fused filters as our compact CNN model without any sparse constraint. Our DCFF not only advances in its simple implementation, but also shows superior ability to derive more compact models with better classification performance when compared to many recent competitors.

\section*{Acknowledgments}
This work was supported by the National Science Fund for Distinguished Young Scholars (No. 62025603), the National Natural Science Foundation of China (No. U21B2037, No. 62176222, No. 62176223, No. 62176226, No. 62072386, No. 62072387, No. 62072389, and No. 62002305), Guangdong Basic and Applied Basic Research Foundation (No. 2019B1515120049), and the Natural Science Foundation of Fujian Province of China (No. 2021J01002).

\ifCLASSOPTIONcaptionsoff
  \newpage
\fi




\bibliographystyle{IEEEtran}
\bibliography{main}

%



%

\begin{IEEEbiography}[{\includegraphics[width=1in,height=1.25in,clip,keepaspectratio]{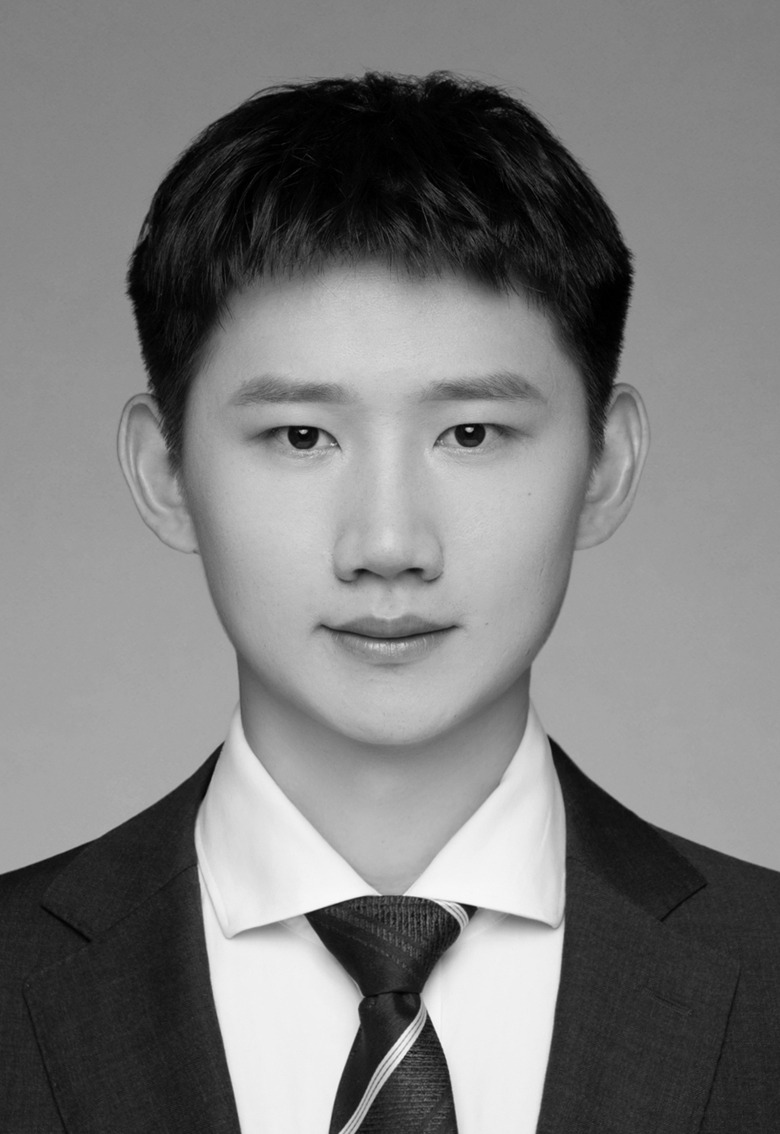}}]{Mingbao Lin} finished his M.S.-Ph.D. study and obtained the Ph.D. degree in intelligence science and technology from Xiamen University, Xiamen, China, in 2022. Earlier, he received the B.S. degree from Fuzhou University, Fuzhou, China, in 2016.

He is currently a senior researcher with the Tencent Youtu Lab, Shanghai, China. His publications on top-tier conferences/journals include IEEE TPAMI, IJCV, IEEE TIP, IEEE TNNLS, CVPR, NeurIPS, AAAI, IJCAI, ACM MM and so on. His current research interest is to develop efficient vision model, as well as information retrieval.
\end{IEEEbiography}

\begin{IEEEbiography}[{\includegraphics[width=1in,height=1.25in,clip,keepaspectratio]{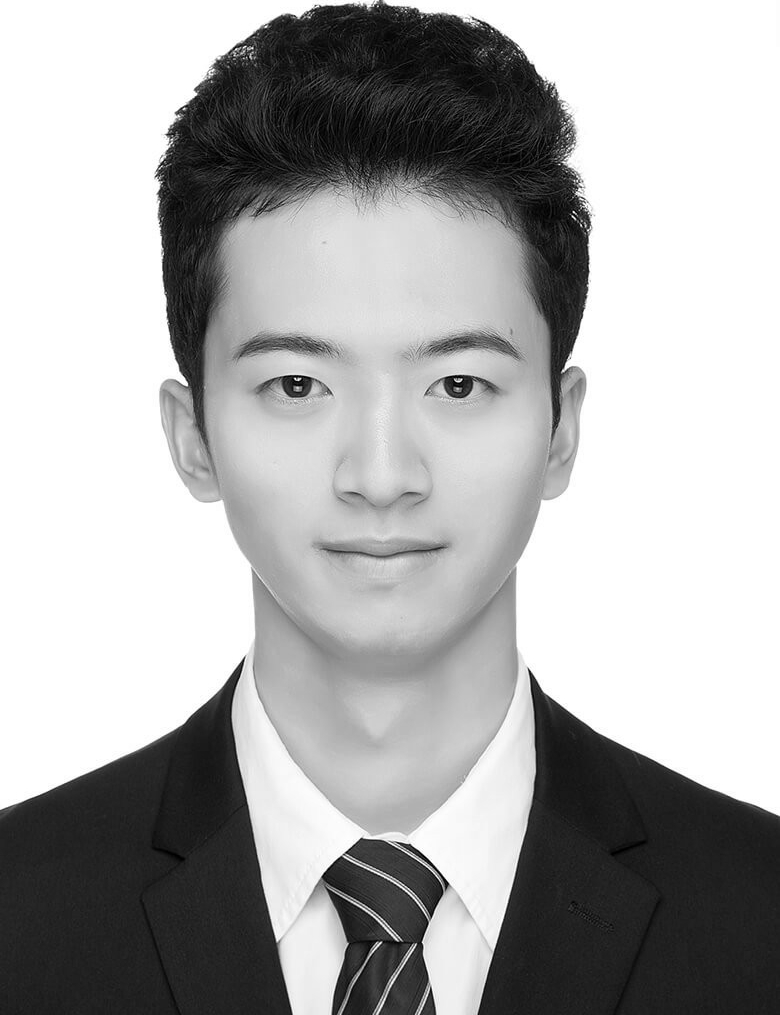}}]{Bohong Chen}
is currently pursuing the B.S. degree with Xiamen University, China. His research interests include computer vision, and neural network compression and acceleration.
\end{IEEEbiography}


\begin{IEEEbiography}[{\includegraphics[width=1in,height=1.25in,clip,keepaspectratio]{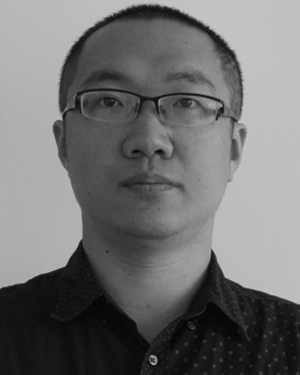}}]{Fei Chao}
(Member, IEEE) received the B.Sc. degree in mechanical engineering from the Fuzhou University, Fuzhou, China, in 2004, the M.Sc. degree with distinction in computer science from the University of Wales, Aberystwyth, U.K., in 2005, and the Ph.D. degree in robotics from the Aberystwyth University, Wales, U.K., in 2009.

He is currently an Associate Professor with the School of Informatics, Xiamen University, Xiamen, China. He has authored/co-authored more than 50 peer-reviewed journal and conference papers. His current research interests include developmental robotics, machine learning, and optimization algorithms.
\end{IEEEbiography}

\begin{IEEEbiography}[{\includegraphics[width=1in,height=1.25in,clip,keepaspectratio]{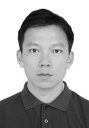}}]{Rongrong Ji}
(Senior Member, IEEE) is a Nanqiang Distinguished Professor at Xiamen University, the Deputy Director of the Office of Science and Technology at Xiamen University, and the Director of Media Analytics and Computing Lab. He was awarded as the National Science Foundation for Excellent Young Scholars (2014), the National Ten Thousand Plan for Young Top Talents (2017), and the National Science Foundation for Distinguished Young Scholars (2020). His research falls in the field of computer vision, multimedia analysis, and machine learning. He has published 50+ papers in ACM/IEEE Transactions, including TPAMI and IJCV, and 100+ full papers on top-tier conferences, such as CVPR and NeurIPS. His publications have got over 10K citations in Google Scholar. He was the recipient of the Best Paper Award of ACM Multimedia 2011. He has served as Area Chairs in top-tier conferences such as CVPR and ACM Multimedia. He is also an Advisory Member for Artificial Intelligence Construction in the Electronic Information Education Committee of the National Ministry of Education.
\end{IEEEbiography}

\end{document}